\documentclass{article}
\usepackage{spconf,amsmath,amssymb,graphicx,hyperref}
\usepackage{booktabs}

\title{Denoising of Two-Phase Optically Sectioned Structured Illumination Reconstructions Using Encoder-Decoder Networks}

\name{Allison Davis, Yezhi Shen, Xiaoyu Ji, Fengqing Zhu \thanks{This work is supported by the Office of Naval Research under the Multidisciplinary University Research Initiatives (MURI) Grant Award N000142412545.}}
\address{Author Affiliation(s)}
\address{Elmore Family School of Electrical and Computer Engineering\\
    Purdue University\\
	West Lafayette, Indiana, USA}

\begin{document}
%
\maketitle
%

\begin{abstract}
Structured illumination (SI) enhances image resolution and contrast by projecting patterned light onto a sample. In two-phase optical-sectioning SI (OS-SI), reduced acquisition time introduces residual artifacts that conventional denoising struggles to suppress. Deep learning offers an alternative to traditional methods; however, supervised training is limited by the lack of clean, optically sectioned ground-truth data. We investigate encoder-decoder networks for artifact reduction in two-phase OS-SI, using synthetic training pairs formed by applying real artifact fields to synthetic images. An asymmetrical denoising autoencoder (DAE) and a U-Net are trained on the synthetic data, then evaluated on real OS-SI images. Both networks improve image clarity, with each excelling against different artifact types. These results demonstrate that synthetic training enables supervised denoising of OS-SI images and highlight the potential of encoder-decoder networks to streamline reconstruction workflows.
\end{abstract}


\section{Introduction}
\label{sec:intro}
Structured illumination (SI) is an imaging technique that enhances visualization by using patterned light to modulate samples, enabling both optical sectioning and super-resolution microscopy. In optical sectioning, SI suppresses out-of-focus light, whereas in super-resolution microscopy, it shifts high-frequency information into the microscope's passband \cite{kristensson_2012}. Both approaches require multiple phase-shifted sub-images that are demodulated into a single reconstruction, followed by targeted post-processing to mitigate residual artifacts.

\begin{figure*}[t]
  \centering
  \includegraphics[width=1.0\textwidth]{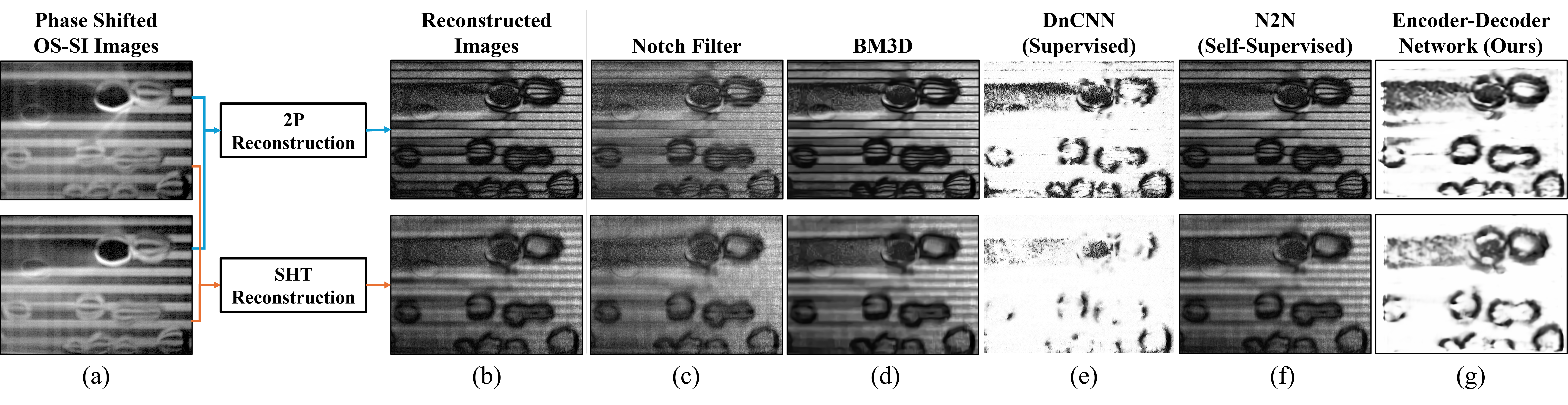}
  \caption{Comparison of denoising techniques on two reconstruction methods. (a) Consecutive images used in reconstruction; (b) 2P (top) and SHT (bottom) reconstructions. (c)-(f) Denoised outputs from notch filtering, BM3D, DnCNN (supervised on synthetic data), and N2N (self-supervised on real OS-SI images). (g) Results from our DAE encoder-decoder network.}
  \label{fig:overview}
\end{figure*}

Neil et al. introduced SI for optical sectioning using three phase-shifted sub-images \cite{neil_1997j}. While this configuration cleanly separates modulated and unmodulated components, it increases acquisition time, limiting its use for dynamic applications \cite{kristensson_2014j}. Two-phase methods reduce acquisition but cannot fully separate modulated components, resulting in reconstruction artifacts \cite{kristensson_2014j}. These artifacts vary across images due to experimental factors such as illumination shadowing or optical system changes, making consistent denoising difficult. Fig.~\ref{fig:overview} shows examples of these artifacts, which are poorly mitigated by traditional methods such as notch filters and advanced denoisers like BM3D \cite{dabov_2007j}.

Deep learning offers an alternative by learning to suppress artifacts directly from data. CNNs such as DnCNN and FFDNet perform well on Gaussian noise but struggle with complex or spatially varying noise \cite{zhang_2017j, zhang_2018j}, as shown in Fig.~\ref{fig:overview}. Encoder-decoder networks address this by compressing input data into a latent space of essential features, then reconstructing the target image. Vincent et al. first introduced the use of denoising autoencoders for robust feature extraction \cite{vincent_2008c}, which was later extended by architectures such as U-Net with skip connections to preserve spatial detail \cite{ronneberger_2015c}. While supervised training typically requires noisy–clean pairs, such data are difficult to obtain. Self-supervised methods such as Noise2Noise (N2N) and Noise2Void showed that clean targets are not strictly necessary \cite{lehtinen_2018c, krull_2019c}. However, these approaches assume i.i.d Gaussian noise and break down when the noise is spatially correlated and structured. StructN2V addresses this with shaped blind masks, but struggles when noise overlaps with real image features \cite{broaddus_2020c}. These limitations highlight the need for domain-specific strategies tailored to OS-SI artifacts.

In this work, we apply encoder-decoder networks to denoise two-phase OS-SI reconstructions. Since clean ground-truth OS-SI data are unavailable, we generate synthetic datasets corrupted with real SI reconstruction artifacts, enabling supervised training with clean-noisy image pairs. The main contributions of this work are as follows:

\begin{itemize}
    \item We propose a synthetic data strategy to generate realistic artifact-corrupted OS-SI training pairs.
    \item We demonstrate the trained encoder-decoder networks can successfully denoise real OS-SI reconstructions.
\end{itemize}

\section{Description of Structured Illumination Reconstruction}
\label{sec:SI_reconstruction}
As introduced in Section \ref{sec:intro}, OS-SI typically uses three phase-shifted sub-images. Each sub-image, $I_n$, is expressed as $I_n(x,y) = I_D+ I_F \cdot \sin(2\pi vy + \Phi_n)$, where $I_D$ is the defocused wide-field component, $I_F$ the in-focus optically sectioned image, and $v$, $\Phi_n$ the spatial frequency and phase of the illumination pattern, respectively \cite{berrocal_2008j}. 

For the three-phase case with $\Phi_n = 0, {2\pi}/{3}, {4\pi}/{3}$, the optically sectioned image is reconstructed as:
\begin{equation}
    I_F = \frac{\sqrt{2}}{3}\sqrt{(I_0 - I_1)^2 + (I_0 - I_2)^2 + (I_1 - I_2)^2}
\end{equation}

In the two-phase case ($\Phi_n = 0, \pi$), the reconstruction simplifies to \cite{berrocal_2016c}:
\begin{equation}
\label{eq:2p_slipi}
    I_F = \frac{\sqrt{2}}{2} \cdot \sqrt{(I_0 - I_1)^2}
\end{equation}

The simplification reduces acquisition time but introduces residual line artifacts due to the missing information normally provided by the third phase-shifted image, requiring Fourier filtering \cite{berrocal_2016c}. We refer to this reconstruction approach as the 2P method.

An alternative two-phase reconstruction method is the Sequence Hilbert Transform (SHT) algorithm \cite{zhou_2015j}. The approach applies a one-dimensional Hilbert transform to the difference image $\Delta I(x, y) = I_1 - I_0$, modeled as \cite{spadaro_2020j}:
\begin{equation}
    \Delta I(x,y) = 2I_F(x,y)\sin(\Phi_0/2)\cos(2\pi vy + \Phi_0 /2)
\end{equation}

The reconstructed image is then obtained by taking the magnitude of the analytic signal:
\begin{equation}
\label{eq:ht_slipi}
    2I_F(x,y) = |\Delta I(x,y) + i\mathrm{HT}_y{\Delta I(x,y)}|
\end{equation} 

However, SHT assumes consistent illumination intensity between the two sub-images and across the field of view. Reconstruction artifacts become prevalent when these conditions are not met. We demonstrate our denoising approach on OS-SI images from both two-phase reconstruction methods.

\section{Method}
\label{sec:method}
Our denoising strategy builds upon the two-phase SI reconstructions in Section \ref{sec:SI_reconstruction}. We train encoder–decoder networks on a synthetic dataset to suppress reconstruction artifacts while preserving structural detail in real images.

\subsection{Model Architectures}
\label{ssec:architectures}
We evaluate two architectures: an asymmetrical denoising autoencoder (DAE) and a U-Net, shown in Fig.~\ref{fig:architectures}. DAEs are computationally lightweight but lose high-frequency details, whereas U-Nets use skip connections to preserve spatial structure, but at the cost of higher model complexity.

\subsubsection{Asymmetrical Denoising Autoencoder}
\label{sssec:dae_arch}
The asymmetrical DAE (Fig.~\ref{fig:architectures}a) is designed with fewer layers in the encoder than the decoder. This allows spatial information to be compressed to the latent representation quickly, followed by gradual upsampling in the decoder to mitigate upsampling artifacts. The encoder employs three levels of convolution layers, each with ReLU activation.

For reconstruction, the decoder uses upsampling convolution blocks, rather than 2D transposed convolutions, which are prone to checkerboard artifacts from uneven kernel overlap \cite{odena_2016j}. Each block first performs bilinear interpolation to increase spatial resolution, followed by a convolutional layer to reduce the number of feature maps. Six such blocks are used to gradually upsample and reconstruct the clean image. As with the encoder, ReLU activation is used at each layer. At the final output, sigmoid activation is used to constrain the output to a normalized range.

\subsubsection{Denoising U-Net}
\label{sssec:unet_arch}
\begin{figure}[htb]
  \centering
  \includegraphics[width=0.7\linewidth]{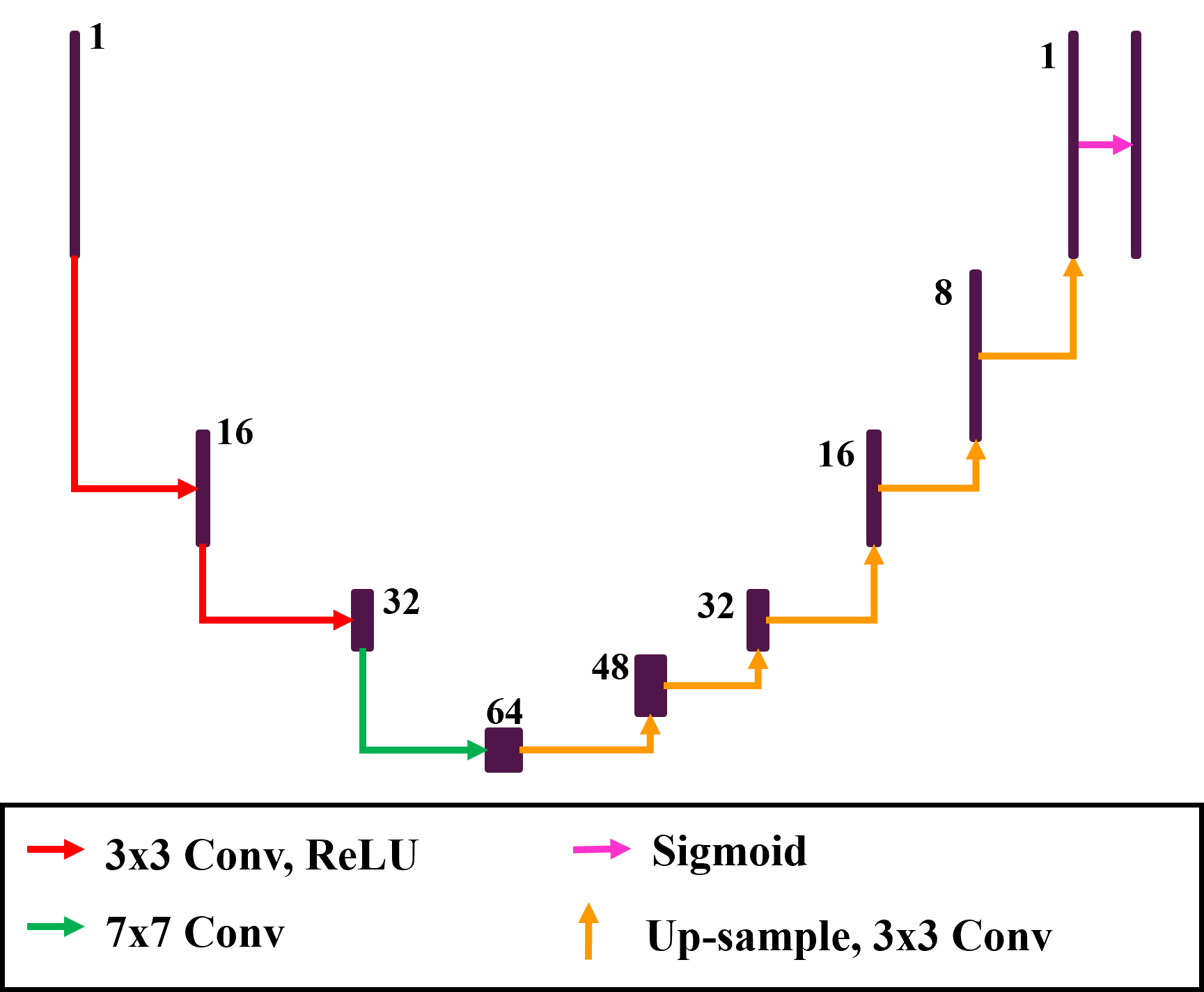}
  \centerline{(a)}\medskip
  
  \includegraphics[width=0.7\linewidth]{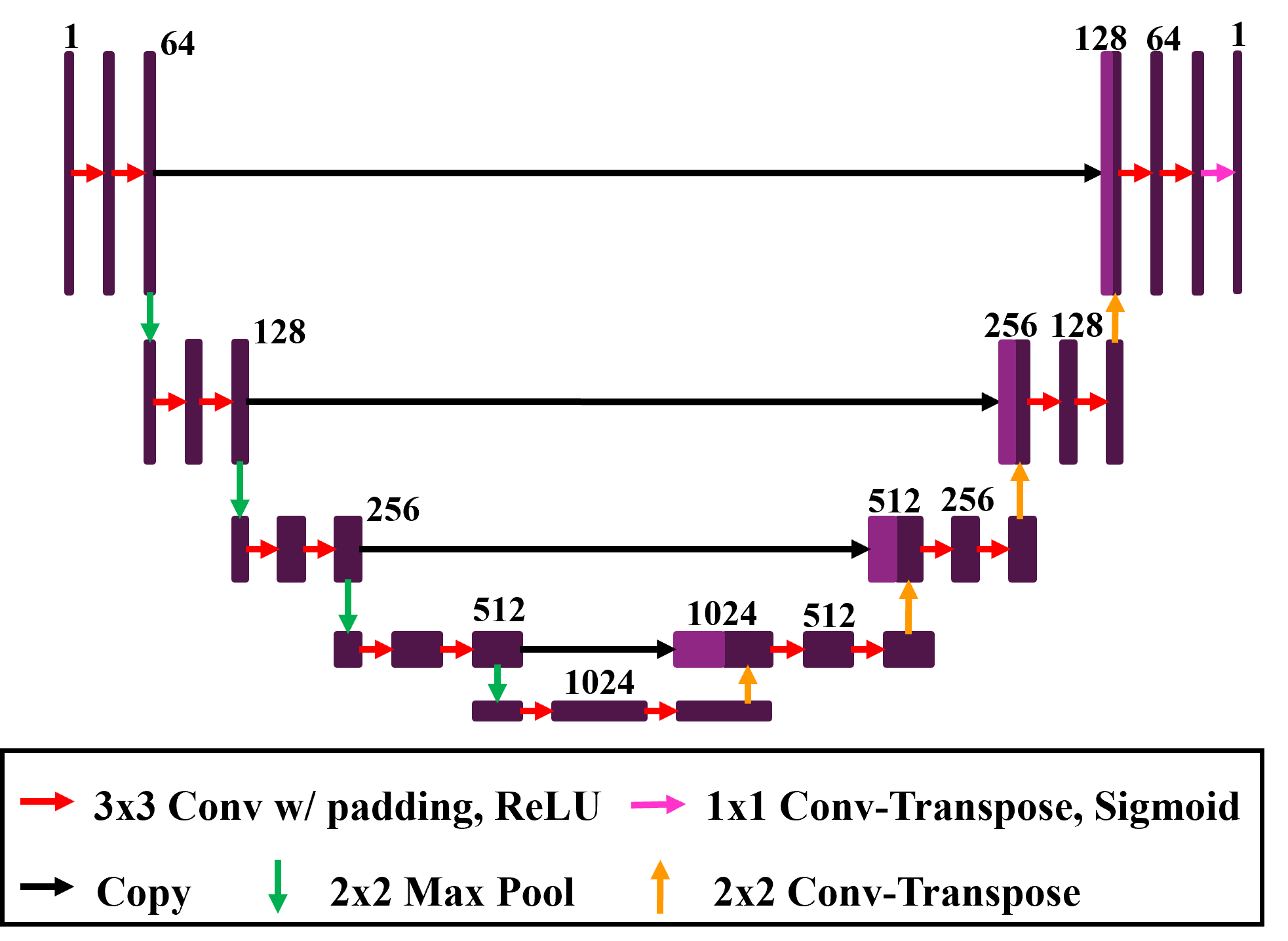}
  \centerline{(b)}\medskip

  \caption{Network architectures used for artifact reduction. (a) Asymmetrical DAE, consisting of a downsampling encoder, a latent representation, and an upsampling decoder. (b) U-Net architecture with symmetric skip connections, which better preserves spatial detail during reconstruction.}
  \label{fig:architectures}
\end{figure}

The denoising U-Net (Fig.~\ref{fig:architectures}b) follows the same architecture introduced by Ronneberger et al., with the addition of batch normalization to stabilize training \cite{ronneberger_2015c}. The encoder consists of four downsampling blocks, each composed of two convolutions with ReLU activations and batch normalization, followed by a max pooling layer.

The decoder mirrors the encoder with four upsampling blocks. Each begins with a transposed convolution that doubles the spatial resolution and halves the channel depth. The result is concatenated with the corresponding encoder feature map via a skip connection. At the final layer, a convolution reduces the output to a single channel, with sigmoid activation to normalize the intensity values.

\subsection{Datasets}
\label{ssec:datasets}
An important component of this work is the use of a synthetic dataset to assist in denoising reconstructed two-phase SI data. Since denoising networks benefit from training data with structures resembling the test domain, we generate synthetic bubble images that resemble the real data \cite{wang_2023j}. These images are then corrupted using reconstruction artifacts from real SI images.

\subsubsection{Training Data: Synthetic Bubbles Dataset}
\label{sssec:synth_bubbles}
We follow the method presented in \cite{gong_2022j} to generate individual synthetic bubbles. Bubble boundaries are approximated by mapping a circle in the complex plane using the Joukowski transform. Scaling, rotation, and skew control the size and shape, while a depth map projects 3D geometry into 2D intensity images. Full images are formed by randomly placing bubbles across the frame, with a bias toward the bottom. Each bubble is randomly scaled, and variation in size, placement, and number increases the dataset's diversity.

\subsubsection{Test Data: Structured-Illumination Pool-Boiling Dataset}
\label{sssec:slipi_data}
The real dataset of interest consists of images of pool-boiling bubbles illuminated with a structured light sheet. This dataset presents unique challenges due to the presence of shadowing from the bubbles, light scattering off bubble surfaces, and a gradual illumination decay across the field. Additionally, pool-boiling is inherently dynamic, meaning the scene evolves between consecutive sub-image acquisitions. As a result, small differences in bubble position or shape contribute to degradations in the reconstructions. Fig.~\ref{fig:overview} shows examples from this dataset, highlighting two consecutive frames and the resulting reconstructions.

\subsubsection{Noising of Training Data}
\label{sssec:noising}
To generate noisy training pairs, we apply real artifact fields obtained from SI reconstructions of bubble-free sub-images. Given a clean synthetic bubble image, $I(x,y)$ and an artifact field $N(x,y)$, the corrupted image is modeled as:
\begin{equation}
    I_N(x, y) = I(x,y)\cdot N(x,y)
\end{equation}

\begin{figure}[htb]
\begin{minipage}[b]{0.48\linewidth}
  \centering
  \centerline{\includegraphics[width=4.2 cm]{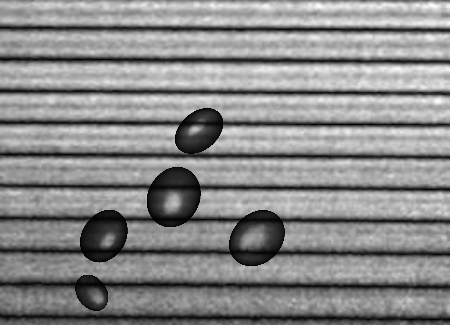}}
  \centerline{(a)}\medskip
\end{minipage}
\hfill
\begin{minipage}[b]{0.48\linewidth}
  \centering
  \centerline{\includegraphics[width=4.2 cm]{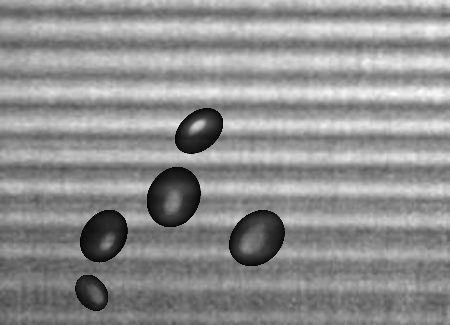}}
  \centerline{(b)}\medskip
\end{minipage}
\caption{Synthetic-bubble images with two-phase OS-SI reconstruction artifacts. (a) Residuals from 2P reconstruction. (b) Residuals derived from SHT reconstruction.}
\label{fig:noisy_images}
\end{figure}

To increase training diversity, we apply random spatial crops and scalings to the artifact fields, varying spatial frequency and illumination. A total of 35,000 images are generated, split evenly between 2P and SHT artifacts. Fig.~\ref{fig:noisy_images} shows the generated synthetic bubbles, each corrupted with a different type of reconstruction artifact.

\section{Experiments}
\label{sec:experiments}
Both models were trained for 50 epochs (batch size = 16, $\alpha=10^{-4}$) on 28{,}000 synthetic images and evaluated on 7{,}000 held-out test images, with a DnCNN trained on the same dataset for comparison. Table~\ref{tab:psnr_results} reports average PSNR on the synthetic bubble test set, separated by reconstruction method (2P and SHT). While DnCNN outperforms the asymmetrical DAE, the U-Net achieves the highest performance by a wide margin. On example test images (Fig.~\ref{fig:noisy_images}), all trained networks substantially outperform conventional denoisers, with notch filtering (9.82 dB, 12.92 dB) and BM3D (6.94 dB, 7.98 dB) producing far lower PSNR.

\begin{table}[htb]
\centering
\caption{Average PSNR (dB) on the synthetic bubble test set, evaluated for 2P and SHT reconstructions using DnCNN and encoder–decoder networks (DAE and U-Net).}
\label{tab:psnr_results}
\begin{tabular}{lccc}
\toprule
        & DnCNN 
        & DAE 
        & U-Net \\
\midrule
2P      & 38.04  & 22.78  & 51.86 \\
SHT     & 40.44  & 22.82  & 54.74 \\
\bottomrule
\end{tabular}
\end{table}

\begin{figure}[htb]
\centering
\begin{minipage}[b]{0.48\textwidth}
  \centering
  \includegraphics[width=\textwidth]{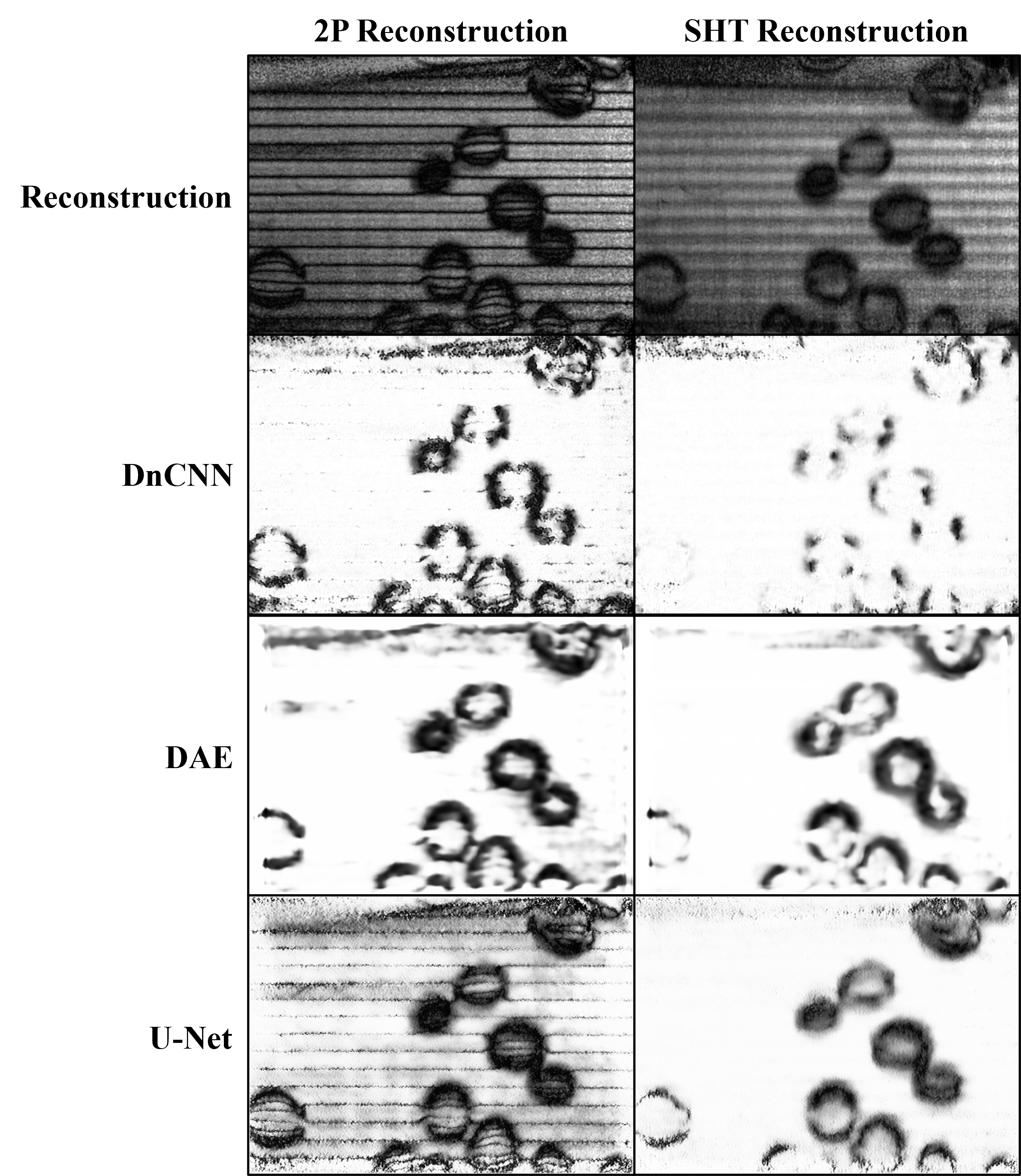}
\end{minipage}
\caption{Qualitative denoising on two samples of real two-phase OS-SI pool-boiling images. The rows show 2P and SHT reconstructions (top), denoised reconstructions using the asymmetrical DAE (middle), and denoised reconstructions using the U-Net (bottom).}
\label{fig:results}
\end{figure}

The trained models were next applied to real pool-boiling SI data reconstructed with both 2P and SHT methods. Fig.~\ref{fig:results} shows representative results. The top row presents reconstructions with residual artifacts, followed by denoised results from DnCNN, the asymmetrical DAE, and the U-Net. Because no ground-truth clean images are available, evaluation is conducted qualitatively.

For both reconstruction types, the DAE and U-Net outperform DnCNN in preserving bubble structures. In the SHT case, the DnCNN loses nearly all bubble information, despite outperforming the DAE quantitatively on the synthetic test set (Table~\ref{tab:psnr_results}). For 2P reconstructions, the DAE more effectively suppresses line artifacts than the U-Net, particularly in background regions and within bubbles, but at the expense of increased edge blurring. This difference suggests that the DAE prioritizes artifact removal, whereas the U-Net emphasizes structural preservation. In contrast, the U-Net shows stronger performance on SHT reconstructions, where artifacts manifest as non-uniform illumination. Skip connections help preserve bubble geometry while reducing background variation. The DAE also reduces SHT artifacts but tends to oversmooth and distort bubble edges, whereas the U-Net generally preserves boundary continuity. These results suggest that the U-Net is particularly well-suited for denoising reconstruction types in which artifacts overlap less with fine structural details.

 Overall, both encoder–decoder architectures improve visual clarity over noisy reconstructions, with complementary strengths. The asymmetrical DAE excels at removing structured line residuals, while the U-Net better handles illumination artifacts and preserves edge details. The encoder–decoder networks are better suited than traditional filters and other learning-based models for reducing reconstruction artifacts while preserving essential object features in real OS-SI images.

\section{Conclusion}
\label{sec:conclusion}
We investigated encoder-decoder networks for denoising two-phase OS-SI reconstructions, in which reduced image acquisition introduces residual artifacts. To overcome the lack of clean ground-truth SI data, we generated synthetic bubble images with applied artifacts from 2P and SHT reconstructions and trained an asymmetrical DAE and a U-Net for supervised denoising. On real pool-boiling data, both models improved image clarity: the DAE achieved stronger suppression of structured line artifacts from 2P reconstruction, whereas the U-Net better mitigated non-uniform illumination from SHT reconstructions and preserved bubble edges. Although synthetic data enables effective denoising of OS-SI, performance is limited by dataset scope. Future work should expand the synthetic dataset and explore domain adaptation to account for motion, shadowing, and scattering. Overall, deep learning presents a promising path toward simplifying OS-SI workflows by reducing reliance on targeted post-processing.

\newpage
\bibliographystyle{IEEEtran}
\bibliography{ICASSP2026}   
\end{document}